\pdfoutput=1

\documentclass[11pt]{article}

\usepackage[final]{acl}

\usepackage{times}
\usepackage{latexsym}
\usepackage[english]{babel}
\usepackage{paralist, tabularx, booktabs}
\usepackage[T1]{fontenc}

\newcommand\setalign{4pt}
\usepackage[utf8]{inputenc}

\usepackage{microtype}
\usepackage{array}
\usepackage{multirow}
\usepackage{graphicx}
\usepackage{amsmath}
\usepackage{tabularray}
\usepackage{bbding}
\usepackage{paralist}
\usepackage{float}

\usepackage{inconsolata}

\title{On Creating an English-Thai Code-switched Machine Translation\\in Medical Domain}

\author{
Parinthapat Pengpun\textsuperscript{1,†},
Krittamate Tiankanon\textsuperscript{2,†},
Amrest Chinkamol\textsuperscript{2,3}, \\
\textbf{Jiramet Kinchagawat}\textsuperscript{2},
\textbf{Pitchaya Chairuengjitjaras}\textsuperscript{2,4},
\textbf{Pasit Supholkhan}\textsuperscript{5},\\
\textbf{Pubordee Aussavavirojekul}\textsuperscript{2},
\textbf{Chiraphat Boonnag}\textsuperscript{7},
\textbf{Kanyakorn Veerakanjana}\textsuperscript{2,5,6},\\
\textbf{Hirunkul Phimsiri}\textsuperscript{4},
\textbf{Boonthicha Sae-jia}\textsuperscript{4},
\textbf{Nattawach Sataudom}\textsuperscript{2},\\
\textbf{Piyalitt Ittichaiwong}\textsuperscript{2,5,6,*},
\textbf{Peerat Limkonchotiwat}\textsuperscript{3,*} \\
\textsuperscript{1}Bangkok Christian International School,
\textsuperscript{2}PreceptorAI team, CARIVA Thailand, \\
\textsuperscript{3}Vidyasirimedhi Institute of Science and Technology,
\textsuperscript{4}Chulalongkorn University,\\
\textsuperscript{5}Mahidol University,
\textsuperscript{6}King's College London,
\textsuperscript{7}Independent researcher\\
\textsuperscript{†}Equal contribution,
\textsuperscript{*}Corresponding authors\\
\texttt{piyalitt.itt@preceptorai.tech},
\texttt{peerat.l\_s19@vistec.ac.th}
}

\begin{document}
\maketitle
\begin{abstract}
Machine translation (MT) in the medical domain plays a pivotal role in enhancing healthcare quality and disseminating medical knowledge. Despite advancements in English-Thai MT technology, common MT approaches often underperform in the medical field due to their inability to precisely translate medical terminologies. Our research prioritizes not merely improving translation accuracy but also maintaining medical terminology in English within the translated text through code-switched (CS) translation. We developed a method to produce CS medical translation data, fine-tuned a CS translation model with this data, and evaluated its performance against strong baselines, such as Google Neural Machine Translation (NMT) and GPT-3.5/GPT-4. Our model demonstrated competitive performance in automatic metrics and was highly favored in human preference evaluations. Our evaluation result also shows that medical professionals significantly prefer CS translations that maintain critical English terms accurately, even if it slightly compromises fluency. Our code and test set are publicly available \url{https://github.com/preceptorai-org/NLLB_CS_EM_NLP2024}.
\end{abstract}

\section{Introduction}
Medical-domain machine translation (MT) serves as a critical component in enhancing healthcare quality and disseminating medical knowledge. By providing accurate translations of medical research publications, MT enables local medical professionals without English proficiency to overcome the linguistic barrier and have access to more medical academic resources, which are predominantly written in English \cite{pecina2014adaptation, mclean2013poor}. This accessibility is crucial for facilitating continuing medical education, which has been shown to be an effective strategy for healthcare professionals to enhance care quality and patient outcomes \cite{bloom2005effects, Randhawa2013}.

\begin{figure}[t]
    \centering
    \includegraphics[width=7cm]{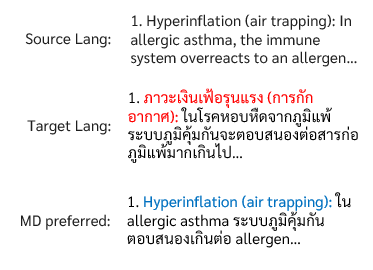}
    \vspace{-6mm}
    \caption{Example of how Google NMT alters the meaning of the sentence when translating from the source language (English) to the target language (Thai) and how it compares with the translations that Medical Doctors (MDs) prefer. \textcolor{blue}{``Hyperinflation''} (abnormal increase of lung volume) is translated into \textcolor{red}{Hyperinflation} in economic context; \textcolor{blue}{``air trapping''} (retention of air in the lungs distal to an obstruction) is translated into \textcolor{red}{``air quarantine''}.}
    \label{img:example-cs}
    \vspace{-9mm}
\end{figure}

Despite the research in the English-Thai MT field, most of the common MT techniques are not yet suitable for the medical domain due to the need for precise translation of medical terminology. Translating medical terminology accurately is challenging due to the lack of equivalent Thai terms for some English medical keywords. Thus, it is understandable why common techniques of machine translation in the medical domain MT --- such as Google NMT, No Language Left Behind (NLLB)~\cite{nllb_team_no_2022}, GPT-4~\cite{achiam2023gpt}, or Gemini-Pro~\cite{team2023gemini} --- cannot translate medical keywords precisely, as shown in Figure~\ref{img:example-cs}. Previous studies have aimed to improve translation accuracy through terminology integration \cite{nieminen_opus-cat_2023, semenov_findings_2023}, yet their application to the Thai language and specifically in the medical field remains limited. 
%


Rather than focusing on enhancing the accuracy of terminology translation, our objective is to preserve medical terms in their original English form within the translated output, thus removing the need to handle terminology translation. This strategy is characterized as \textbf{code-switching (CS)}, deviating from conventional monolingual translation practices. Apart from reducing the task's complexity, framing this problem as CS is also preferred by Medical Doctors (MDs). A previous study from the Thai COVID-19 administrative unit~\cite{Toomaneejinda2022} suggests that medical professionals prefer retaining medical and technical keywords in English, with the rest of the translation in Thai, to avoid potential translation inaccuracies. Other studies ~\cite{Alqurashi2022, IndahCodeSwitching2019, Wood2018Departing, RodriguezTembras2016} also suggest that this phenomenon exists in other languages as well, including Arabic, Javanese, and Spanish. As shown in Figure~\ref{img:example-cs}, keeping medical terms in English preserves the original meaning and is preferred by MDs.

However, CS datasets are usually scarce, preventing researchers from developing a CS translator. Several initiatives have been made to address this issue. For example, the LinCE dataset~\cite{aguilar_lince_2020} is one of the publicly available CS datasets focused on the general domain created to mitigate this problem. Additionally, various studies have focused on enhancing CS dataset efficiency through augmentation techniques~\cite{gowda_checks_2022, sugiyama_data_2019, menacer_machine_2019}, pre-training techniques~\cite{Yang2020CSP,iyer-etal-2023-code}, synthetic CS dataset generation techniques~\cite{tarunesh_machine_2021,appicharla-etal-2021-iitp,xu-yvon-2021-traducir}. However, these previous works did not focus on medical texts. Furthermore, the adaptation of such research to the Thai language context remains limited. This, in turn, leads to a significant scarcity of the English-Thai CS translation dataset, especially in the medical domain, as shown in Table~\ref{tab:dataset_comparison}.

In this work, we aim to achieve two objectives: (i) address the data scarcity in medical-domain English-Thai MT and (ii) validate our hypothesis that doctors prefer CS translations to monolingual ones in medical contexts. 
To achieve the first objective, we create a new English-Thai CS dataset for the medical domain. 
%
Our process begins by generating initial CS translations (pseudo-CS) of English medical texts using a widely available monolingual translator. During this translation process, we apply a keyword masking algorithm to preserve key medical terms. 
%
We then hire an annotator to post-process and clean a portion of the generated translations, as opposed to doing the whole translation process, to save both time and resources.
%

To achieve our second objective, we conduct comprehensive evaluations using both traditional MT metrics and MD evaluations to confirm our hypothesis.
This involves evaluating 52 models, including an off-the-shelf translator, large language models (LLMs), and our fine-tuned CS models based on NLLB. 
Furthermore, we assess the translations for factual accuracy and MD preference by having MDs directly rate them and by distributing preference ranking questionnaires, respectively. 
%
%
Our findings reveal that our fine-tuned CS model based on NLLB is preferred by MDs due to its factual accuracy, even though it achieves a lower score in traditional metrics when compared to off-the-shelf translators like Google NMT.
These results indicate that traditional MT metrics are inadequate for evaluating medical-domain translations and that MDs prefer CS translations over monolingual ones.

To summarize, our key contributions are:
\begin{compactitem}[\hspace{\setalign}•]
    \item We propose the first benchmark dataset specifically designed for medical English-Thai CS translation.
    \item We develop the first open-source model tailored for medical English-Thai CS translation, which is preferred by MDs over Google NMT and GPT-3.5 systems.
    \item We present a comprehensive evaluation of various models on our benchmark, including 52 models, 8 metrics, and 27 MD evaluators. These results reveal a misalignment between traditional MT metrics and the judgments of medical professionals, and underscore the preference of MDs for CS translations.
    \item Our code, test set, and translation models are publicly available at \url{https://github.com/preceptorai-org/NLLB_CS_EM_NLP2024}.
\end{compactitem}

\section{Related Works}
\label{sec:related_works}

\begin{table*}[!ht]
    \centering
    
    \caption{Comparison of English-to-Thai translation datasets. Given \#Samples is the number of samples, \#Sentences is the number of sentences, and \#English Tokens is the count of English tokens within all of the target translations. The Ratio En:All column reflects the proportion of English token usage compared to other languages within the target translations. The CS column calculation is based on the percentage of English tokens in target translations.}
    \scalebox{0.80}{
        \begin{tabular}{lrrrrcc}
            \hline
            \textbf{Dataset} & \rotatebox{45}{\textbf{\#Samples}} & \rotatebox{45}{\textbf{\#Sentences}} & \rotatebox{45}{\parbox{3cm}{\textbf{\#English Tokens}}} & \rotatebox{45}{\textbf{Ratio En:All}} & \rotatebox{45}{\textbf{Domain}} & \rotatebox{45}{\textbf{CS?}} \\
            \midrule
            FLORES-200 & 2,009 & 3,251 & 442 & 1.3\% & Wikidata & \XSolidBrush \\
            Thai US Embassy & 615 & 9,303 & 11,176 & 4.7\% & News & \XSolidBrush \\
            SCB\_MT\_EN-TH\_2020 & 1,001,752 & 1,084,328 & 8,124,662 & 1.4\% & General & \XSolidBrush \\
            Our Pseudo-CS & 63,982 & 188,037 & 640,951 & 16.1\% & Medical & \CheckmarkBold \ \\ \hline
        \end{tabular}}
    \label{tab:dataset_comparison}
    \vspace{-3mm}
\end{table*}

\subsection{Neural Machine Translation (NMT)}

NMT has gained prominence in both academic and commercial sectors, largely due to advancements in Transformer-based architectures ~\cite{vaswani_attention_2023}. Various models designed for NMT, such as mT5 \cite{xue_mt5_2021}, mBART \cite{liu_multilingual_2020}, OPUS \cite{tiedemann-thottingal-2020-opus}, and NLLB \cite{nllb_team_no_2022}, have been developed. However, as previously mentioned, most of these NMT models cannot perform precise terminology translations, which disqualifies them from the medical domain.

The emergence of Large Language Models (LLMs) has further changed the NMT landscape. LLMs, such as GPT-4, have demonstrated emergent abilities in machine translation, excelling in paragraph-level translations without the need for extensive fine-tuning on large parallel corpora~\cite{wei2022emergent, kocmi-etal-2023-findings}. A few studies~\cite{zhu2023multilingual, robinson_chatgpt_2023, bawden-yvon-2023-investigating} have suggested that LLMs are not yet effective translators, especially in low-resource languages including Thai. Nevertheless, it has been shown that LLMs are proficient at generating CS data for many languages~\cite{yong-etal-2023-prompting}. To the best of our knowledge, no research has comprehensively investigated the performance of LLMs (both closed and open-source) in translating the Thai language, especially in the medical domain.

\subsection{Evaluation Metrics for NMT}

The assessment of Machine Translation (MT) quality is a continually evolving field of research. Several automated metrics have been proposed to measure MT quality through lexical analysis, including BLEU \cite{papineni_bleu_2001}, chrF \cite{popovic_chrf_2015}, METEOR~\cite{banerjee-lavie-2005-meteor}, and Translation Edit Rate \cite{snover_study_nodate}. Furthermore, various neural-network-based metrics have been devised to enhance the measurement of MT quality using neural networks: COMET \cite{rei_comet_2020}, Mask-Language-Modeling Score \cite{zheng_self-supervised_2021}, and BLEURT~\cite{sellam-etal-2020-bleurt}. While these metrics provide effective means to assess translations, several studies have also shown their limitations, indicating that these metrics do not always correlate well with human evaluation~\cite{mathur-etal-2020-tangled, callison-burch-etal-2006-evaluating, Roy2021ReassessingAE}. It still remains unclear whether these metrics align well with the specific use cases of medical-domain MT, where the precise translation of terminology is more important than overall sentence fluency.

Human evaluation is also crucial, especially in a medical context where technically accurate and human-readable translations are necessary. \citet{graham_continuous_2013} attempted to better standardize crowd jurisdictions on with Likert-type continuous rating scales. After that, a band scale was proposed by \cite{menacer_machine_2019, tarunesh_machine_2021} for more consistent qualitative evaluation among human judges. \citet{bai_training_2022, askell_general_2021} introduced the concept of the Elo Rating to benchmark multiple translation systems' performances. Elo Rating allows for a leaderboard-like relative comparison between these systems. All these works provided valuable perspectives on how to conduct human preference evaluations on NMT models. Using these studies as a basis of our human evaluation on translation models, we choose to use an improved Elo-based metric called the Glicko score, which was developed by Glickman to accounts for the uncertainty in Elo-based calculations~\cite{glickman1995comprehensive, glickman1995glicko, glickman_parameter_1999}.


\section{Benchmark Data Collection}

\subsection{English Text collection}
\label{sec:dataset_gen}

Our dataset of English medical texts was collected from an in-house LLM-based application designed to tackle intricate medical questions, with an emphasis on differential diagnosis and multiple-choice problems. The dataset consists of 10,000 medical excerpts, with an additional 250 excerpts reserved for testing purposes.






\subsection{Pseudo-translation Masking and Generation}
\label{sec:keyword_masking_algorithm}

In the absence of an existing CS translator, we adopt a masking approach to create our CS translation dataset. Inspired by the Language Identification (LID) translation pipeline~\cite{ramadurgam_translating_nodate}, this method involves augmenting a standard monolingual translator with a keyword masking strategy. By identifying the important medical keywords and selectively translating the rest of the sentence, this method allows for the retaining of domain-specific terminology after translation. Using this, we establish a \emph{pseudo-CS} translator, which forms the basis of our benchmark dataset.

The overview of the procedure for the Keyword masking algorithm is as follows:
\begin{compactenum}
    \item Use GPT-4 to identify medical keywords in the original English sentence. We specifically chose GPT-4 for its capabilities in Named Entity Recognition (NER) of medical terms (see Appendix~\ref{sec:gpt4-ner-perf} for our evaluation) and its flexibility, which allows us to manually adjust the types of terms to include or exclude in order to mimic medical CS as closely as possible.
    \item Replace each medical keyword with a unique placeholder. This results in an English text where medical terms are masked.
    \item Process the masked English text through an MT model to obtain a masked Thai text. In this text, the non-medical parts are translated, while the placeholders remain untouched.
    \item Substitute the unique placeholder tokens with their original English medical keywords to produce the final Thai-English pseudo-CS translation.
\end{compactenum}

Expanding on Step 3, the masked sentences from the previous step are translated to generate pseudo-CS translations. All English excerpts and their corresponding masked versions are processed through the keyword-masked Google translation system, resulting in Thai pseudo-CS translations. To ensure proper alignment between the English and Thai+English (as in the target translations contain CS between Thai and English)  content, both the original English excerpts and the CS translations are segmented into chunks of fewer than 256 tokens. We then re-validate that the English and Thai texts contain the same number of chunks.
%
%

Regarding the size of our dataset, our dataset size is competitive when compared to existing code-switched datasets. In terms of the number of samples, our dataset has 64K records, while a single language pair within the LinCE has 7k to 67k records. For the total token counts, our dataset has 640K tokens, while a single language pair within LinCE has 33k to 808k tokens.
We split our dataset into 63,982 English-to-Thai CS translation pairs for training and 1,100 translation pairs for the test benchmark.

\subsection{Test dataset Constitution} \label{subsec:test_data}
To ensure the quality of our test set, we employ human annotators to recheck its fluency with the instruction in Figure~\ref{fig:annot-instruct}. After annotation, the dataset goes through an NLP pipeline to correct typos and adjust spacing. Subsequently, it then undergoes another round of validation by MDs to ensure its readability and factual accuracy. The MDs make further corrections to improve the accuracy of the translation chunks compared to their source text. This process ensures that every sentence and CS word is correct as verified by MDs; LLMs only serve to reduce the time spent here.

\subsection{Training Data Procedure} \label{subsec:training}
As mentioned in the previous step, we utilize both human annotators and MDs to assess the quality of our test set. However, applying the same process to the training data would be 64 times more expensive than the test set. To mitigate this issue, we employ data augmentation and filtering techniques to improve the quality of our training dataset.
\begin{compactitem}[\hspace{\setalign}•]
    \item \textbf{Data Augmentation}: Inspired by the back-translation augmentation method~\cite{sugiyama_data_2019}, we prompt Gemini-Pro to rephrase the existing CS translation while retaining a roughly similar CS boundary. The rephrased CS sentences are then back-translated to generate corresponding English sentences, thereby constructing new translation pairs.
    \item \textbf{Filtering}: We filter the training CS translation dataset based on a rough measure of its quality. We use the COMET score metric (which assesses semantic similarity) to estimate the quality of the translation dataset and filter out samples that did not achieve a COMET score of at least 0.6.
\end{compactitem}

\section{Experimental Setup}

\subsection{Baseline Models}
\label{sec:baseline_models}

\noindent
\textbf{Off-the-Shelf Translator} (1 model). We leverage Google NMT as our off-the-shelf translator, utilizing the version released on January 17, 2024.

\noindent
\textbf{Large Language Models} (18 models). This set includes OpenThaiGPT 7B, OpenThaiGPT 13B, Typhoon 7B \cite{pipatanakul_typhoon_2023}, SeaLLM 7B \cite{nguyen_seallms_2023}, Llama2 7B, Llama2 13B \cite{touvron_llama_2023}, Google's Gemini-Pro, GPT 3.5, and GPT 4. Each large language model has two prompt variants: one prompted to generate monolingual translations (denoted as the "MN" variant) and another prompted to generate CS translations (denoted as the "CS" variant). All local LLMs (OpenThaiGPT, Typhoon, SeaLLM, Llama2) are evaluated using bfloat16 precision. The rest are accessed via API calls with default settings and a temperature of 0.1. The GPT models used are based on the 1106-preview version. The Gemini-Pro model is utilized as presented through the API on January 17th, 2024.

\noindent
\textbf{CS Baseline} (6 models): We employ a state-of-the-art language translation model, NLLB. We utilized NLLB 3.3B as a base model and fine-tuned it on six variants of our training dataset (Section~\ref{subsec:training}) as follows:
\begin{compactitem}[\hspace{\setalign}•]
    \item NLLB-1: Initial 64k dataset (64k)
    \item NLLB-2: Augmentation of the 64k dataset (64k)
    \item NLLB-3: Initial dataset plus augmentation of the 64k dataset (128k)
    \item NLLB-4: Filtered 64k dataset (30k)
    \item NLLB-5: Filtered augmentation dataset (40k)
    \item NLLB-6: Filtered 64k dataset plus filtered augmentation dataset (70k)
\end{compactitem}
It is important to note that our augmentation technique, which utilizes an LLM to rephrase translation pairs, likely results in an overall improvement in the COMET score of the augmented dataset. Setting a fixed COMET score threshold for dataset filtration results in the augmented filtered dataset containing more records than the initially filtered dataset.
The exact training configurations are listed in Appendix~\ref{sec:train_config}. 
In addition, the inference is performed using bfloat16 quantization.

\subsection{Evaluation Metrics}
We evaluate 52 translation systems---26 with the masking system and 26 without the masking system during the inference step (see Section~\ref{sec:keyword_masking_algorithm})---using standard machine translation metrics and MD evaluators to further validate our results.

\subsubsection{Machine Metric Evaluation}
We evaluate all our translation models using our MD-annotated test set. The following metrics are employed for evaluation:
\begin{compactitem}[\hspace{\setlength\itemindent}{•}]
    \item Lexical score (BLEU~\cite{papineni_bleu_2001}, chrF~\cite{popovic_chrf_2015}, METEOR~\cite{banerjee-lavie-2005-meteor}).
    \item Translation Edit Rate, which includes Character Error Rate (CER) and Word Error Rate (WER).
    \item Semantic score (COMET~\cite{rei_comet_2020, guerreiro2023xcomet}).
    \item CS boundary F1 Score, inspired by~\cite{sterner_tongueswitcher_2023}. The CS boundary F1 Score is calculated using the common formula, i.e., the harmonic mean of \emph{precision} and \emph{recall}. \emph{Precision} is defined as the proportion of correctly identified English words in the generated translation compared to those in the reference translation. \emph{Recall} is the proportion of English words in the reference translation that are correctly identified in the generated translation.
\end{compactitem}
Details on the implementation of these metrics are provided in Appendix~\ref{sec:eval_metric_impl}.

\subsubsection{Human Evaluation}
Anticipating a lower number of human respondents, we only selected the MD-preferred models for human evaluation.
To ensure that each model is compared against each other at least 30 times within a comprehensive evaluation of 52 models, it would require at least 39,000 data points, or approximately 390 respondents, to achieve a statistically significant result. By selecting only 8--11 models, we can reduce this number to 2,000 data points or 20 respondents.
%
Our methodology is as follows.

\noindent
\textbf{Before human evaluation}
\label{sec:internal_eval}
We assess the \emph{factual} accuracy of translations produced by each model by soliciting evaluations from four medical professionals. These professionals assess each translation's factual correctness using our specific rubric. The evaluation process is outlined as follows:
\begin{compactitem}[\hspace{\setalign}•]
    \item 10 English texts are randomly selected from our test set and translated using 52 different translation systems.
    \item Medical professionals are instructed to individually rate each translation for factual correctness on a scale from 1 to 7, according to a detailed rubric provided in Table~\ref{tab:band_score}. Each medical professional is unaware of the translations' source models, and the sequence of translations they evaluate is randomized to prevent bias.
    \item The score for each model, as rated by an evaluator, is determined by calculating the median of the scores assigned to its translations.
    \item An arithmetic mean of the median scores across all evaluators is then calculated to assign each model its preliminary final score.
\end{compactitem}
Subsequent human evaluations are conducted only on model categories (differentiated by base translation model and usage of keyword masking) that achieved ratings higher than the Google NMT.

\begin{figure}[ht]
\vspace{-1mm}
\centering
\includegraphics[width=6.5cm]{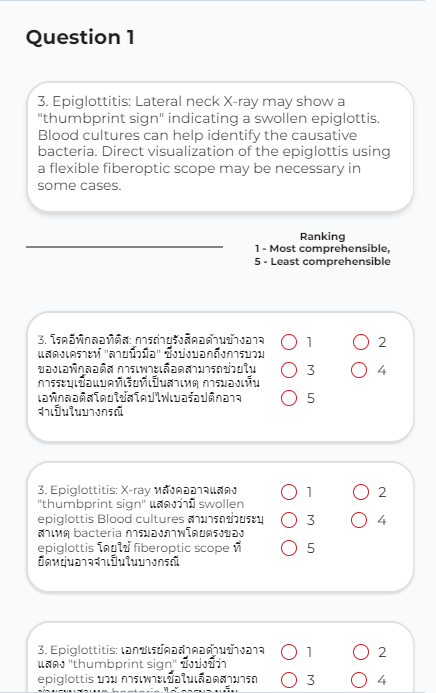}
\caption{Example Questionnaire User Interface}
\vspace{-3mm}
\label{img:questionnaire_sample}
\end{figure}

\noindent
\textbf{Human Preference Evaluation}
\label{sec:human_eval}We perform a human preference evaluation to determine which models are preferred by crowd-sourced medical practitioners, assessing their preference for translations as well as their performance on our human dataset. Note that, in this step, we only ask medical professionals on our chatbot platform to assist in evaluating translations.
\begin{compactitem}[\hspace{\setalign}•]
    \item We evaluate 10 translation models and the human label  based on the previous step. This involves selecting one model from each category identified in the last step.
    \item We design a self-administered, web-based survey using a ranking format to enhance participants' experiences~\cite{Revilla-2020}. Given that ranking five items requires approximately 40 seconds~\cite{Sauro_Atkins_Du_Lewis_2023} and our items consisted of a few sentences, we include ten ranking questions, each estimated to take approximately 1.5 minutes to complete.
    \item For each participant, we randomly sample 10 English texts from our benchmark test set. For each text, we present five versions of the translations, each randomly selected from the list of ``comprehensible'' models along with the human-annotated translation.
    \item Participants are asked to rank each translation sample based on the factual accuracy of the sentence and their preference (as shown in Figure \ref{img:questionnaire_sample}). We specifically instruct them to disregard the proportion of English text retained in the translation (as shown in Figure \ref{fig:questionnaire_sample0}).
    \item Subsequently, we use the preference data from the human evaluation to calculate the \emph{Glicko} Rating, measuring the comparative preference of each model against the others and the human annotator. The initial Glicko rating is set according to the standard, with \(r = 1500\) and \(RD = 350\).
\end{compactitem}
Moreover, we implement a simple filter to monitor each participant's response time to the survey. Participants who completed the survey in less than 5 minutes were flagged as potentially invalid, and their choice ordering was re-examined to confirm the validity of their responses. A row is flagged as an invalid record if the choice order remains nearly identical across questions despite variations in the translation model.

\section{Main Experimental Results}
\label{sec:results}

\begin{table*}[!ht]
    \centering
    \caption{Full Evaluation Result using our dataset. The ``MN'' suffix indicates that the LLM employs a Monolingual translation prompt, whereas the ``CS'' suffix denotes the use of a CS translation prompt. The term ``Mask'' indicates the system's use of a keyword masking algorithm, as described in Section~\ref{sec:keyword_masking_algorithm}. Each NLLB variant is labeled according to the dataset used for training the NLLB, as detailed in Section~\ref{sec:baseline_models}. ``Fact.'' indicates factual accuracy score as described in  Section~\ref{sec:internal_eval}}
    \vspace{-2mm}
    \label{tab:eval_result} %
    \begin{scriptsize}
   \begin{tabular}{l|lllllll|l}
    \hline
Model Variant & CS F1 & BLEU & chrF & CER & WER & COMET & METEOR & Fact.  \\ \hline
    Gemini-Pro-CS & 0.132 & 0.353 & 0.595 & 0.599 & 0.686 & 0.849 & 0.622 & 5.750  \\ 
    Gemini-Pro-MN & 0.110 & 0.352 & 0.599 & 0.426 & 0.526 & \textbf{0.854} & 0.630 & 5.375  \\ 
    Google-NMT & 0.119 & \textbf{0.385} & \textbf{0.617} & \textbf{0.392} & \textbf{0.480} & 0.815 & \textbf{0.650} & 3.125  \\ 
    GPT-3.5-CS & 0.141 & 0.208 & 0.504 & 0.833 & 0.987 & 0.671 & 0.491 & 3.625 \\ 
    GPT-3.5-MN & 0.114 & 0.205 & 0.504 & 0.599 & 0.775 & 0.687 & 0.494 & 3.875  \\ 
    GPT-4-CS & 0.340 & 0.314 & 0.601 & 0.636 & 0.757 & 0.850 & 0.593 & \textbf{6.250}  \\ 
    GPT-4-MN & 0.132 & 0.282 & 0.581 & 0.511 & 0.660 & 0.847 & 0.597 & 5.125 \\ 
    Llama2-13B-CS & 0.058 & 0.012 & 0.189 & 5.226 & 6.104 & 0.153 & 0.129 & 1.125 \\ 
    Llama2-13B-MN & 0.082 & 0.022 & 0.224 & 3.326 & 4.139 & 0.163 & 0.174 & 1.000  \\ 
    Llama2-7B-CS & 0.074 & 0.012 & 0.171 & 5.361 & 6.141 & 0.159 & 0.117 & 0.500  \\ 
    Llama2-7B-MN & 0.086 & 0.015 & 0.197 & 3.852 & 4.761 & 0.162 & 0.143 & 0.500  \\ 
    OpenThaiGPT-13B-CS & 0.039 & 0.094 & 0.446 & 2.343 & 2.538 & 0.394 & 0.388 & 2.125  \\ 
    OpenThaiGPT-13B-MN & 0.036 & 0.094 & 0.465 & 1.978 & 2.208 & 0.425 & 0.396 & 2.375  \\ 
    OpenThaiGPT-7B-CS & 0.045 & 0.046 & 0.308 & 12.620 & 13.224 & 0.310 & 0.237 & 2.875  \\ 
    OpenThaiGPT-7B-MN & 0.027 & 0.068 & 0.344 & 9.954 & 10.223 & 0.369 & 0.282 & 2.750 \\ 
    SeaLLM-7B-CS & 0.035 & 0.017 & 0.242 & 11.678 & 11.165 & 0.235 & 0.188 & 2.000  \\ 
    SeaLLM-7B-MN & 0.076 & 0.032 & 0.329 & 8.340 & 8.259 & 0.321 & 0.259 & 1.705  \\ 
    Typhoon-7B-CS & 0.021 & 0.012 & 0.220 & 18.946 & 18.434 & 0.186 & 0.168 & 1.875 \\ 
    Typhoon-7B-MN & 0.023 & 0.013 & 0.239 & 18.111 & 19.020 & 0.174 & 0.176 & 1.875 \\ 
    NLLB & 0.107 & 0.140 & 0.432 & 0.610 & 0.906 & 0.530 & 0.405 & 2.500 \\ 
    NLLB-1 & \textbf{0.475} & 0.253 & 0.487 & 0.491 & 0.593 & 0.678 & 0.502 & 4.375 \\ 
    NLLB-2 & 0.230 & 0.262 & 0.548 & 0.448 & 0.612 & 0.720 & 0.546 & 3.375  \\ 
    NLLB-3 & 0.380 & 0.257 & 0.520 & 0.472 & 0.604 & 0.702 & 0.521 & 4.000  \\ 
    NLLB-4 & 0.452 & 0.272 & 0.520 & 0.461 & 0.577 & 0.710 & 0.532 & 3.875  \\ 
    NLLB-5 & 0.193 & 0.255 & 0.544 & 0.458 & 0.627 & 0.715 & 0.546 & 3.250 \\ 
    NLLB-6 & 0.286 & 0.264 & 0.539 & 0.456 & 0.606 & 0.711 & 0.541 & 4.000 \\ \hline
    Gemini-Pro-CS + Mask & 0.628 & 0.301 & 0.512 & 0.668 & 0.716 & 0.704 & 0.543 & 5.500 \\ 
    Gemini-Pro-MN + Mask & 0.644 & 0.314 & 0.529 & 0.461 & 0.517 & \textbf{0.726} & 0.562 & 5.750  \\ 
    Google-NMT + Mask & \textbf{0.647} & \textbf{0.327} & \textbf{0.531} & \textbf{0.458} & \textbf{0.509} & 0.656 & \textbf{0.564} & 5.000 \\ 
    GPT-3.5-CS + Mask & 0.574 & 0.212 & 0.463 & 0.839 & 0.953 & 0.631 & 0.468 & 5.250 \\ 
    GPT-3.5-MN + Mask & 0.536 & 0.215 & 0.474 & 0.662 & 0.755 & 0.623 & 0.478 & 5.000  \\ 
    GPT-4-CS + Mask & 0.612 & 0.265 & 0.500 & 0.682 & 0.758 & 0.724 & 0.515 & \textbf{6.000}  \\ 
    GPT-4-MN + Mask & 0.619 & 0.275 & 0.517 & 0.556 & 0.634 & 0.705 & 0.535 & 4.750 \\ 
    Llama2-13B-CS + Mask & 0.052 & 0.011 & 0.164 & 6.050 & 7.205 & 0.142 & 0.110 & 1.000  \\ 
    Llama2-13B-MN + Mask & 0.100 & 0.023 & 0.199 & 4.201 & 5.363 & 0.156 & 0.148 & 0.750  \\ 
    Llama2-7B-CS + Mask & 0.013 & 0.005 & 0.127 & 6.091 & 7.175 & 0.144 & 0.079 & 0.500 \\ 
    Llama2-7B-MN + Mask & 0.024 & 0.008 & 0.150 & 4.188 & 5.712 & 0.161 & 0.101 & 0.750 \\ 
    OpenThaiGPT-13B-CS + Mask & 0.052 & 0.072 & 0.369 & 1.831 & 2.215 & 0.275 & 0.313 & 2.250 \\ 
    OpenThaiGPT-13B-MN + Mask & 0.078 & 0.066 & 0.384 & 2.119 & 2.715 & 0.293 & 0.309 & 1.375 \\ 
    OpenThaiGPT-7B-CS + Mask & 0.043 & 0.038 & 0.266 & 11.545 & 12.430 & 0.226 & 0.202 & 1.250 \\ 
    OpenThaiGPT-7B-MN + Mask & 0.063 & 0.062 & 0.307 & 6.760 & 7.068 & 0.271 & 0.258 & 2.125 \\ 
    SeaLLM-7B-CS + Mask & 0.048 & 0.016 & 0.223 & 10.167 & 9.953 & 0.204 & 0.166 & 1.375 \\ 
    SeaLLM-7B-MN + Mask & 0.163 & 0.033 & 0.306 & 8.119 & 8.009 & 0.259 & 0.240 & 1.625 \\ 
    Typhoon-7B-CS + Mask & 0.080 & 0.011 & 0.199 & 18.283 & 18.291 & 0.170 & 0.147 & 1.875 \\ 
    Typhoon-7B-MN + Mask & 0.113 & 0.010 & 0.218 & 17.891 & 18.786 & 0.172 & 0.150 & 1.750 \\ 
    NLLB + Mask & 0.523 & 0.183 & 0.423 & 0.556 & 0.719 & 0.533 & 0.424 & 4.125 \\ 
    NLLB-1 + Mask & 0.578 & 0.237 & 0.457 & 0.515 & 0.605 & 0.645 & 0.479 & 4.625 \\ 
    NLLB-2 + Mask & 0.637 & 0.240 & 0.475 & 0.506 & 0.612 & 0.644 & 0.489 & 4.750 \\ 
    NLLB-3 + Mask & 0.605 & 0.237 & 0.464 & 0.511 & 0.608 & 0.648 & 0.481 & 5.125  \\ 
    NLLB-4 + Mask & 0.599 & 0.250 & 0.472 & 0.502 & 0.596 & 0.651 & 0.493 & 4.875 \\ 
    NLLB-5 + Mask & 0.642 & 0.242 & 0.478 & 0.504 & 0.609 & 0.645 & 0.493 & 3.625 \\ 
    NLLB-6 + Mask & 0.628 & 0.241 & 0.473 & 0.505 & 0.605 & 0.646 & 0.489 & 4.750 \\ \hline
    \end{tabular}
    \end{scriptsize}
    \vspace{-3mm}
\end{table*}

The full evaluation results for all 52 models on our dataset are presented in Table~\ref{tab:eval_result}. We categorize the results into two groups: (i) traditional machine translation (MT) metrics and (ii) human preference.

\noindent
\textbf{Traditional MT metrics}. We present our two best models: NLLB-1 (initial 64k dataset) and NLLB-4 (filtered 64k dataset) with Mask. These models demonstrate remarkable results among open-source models and achieve competitive results against closed-source models. NLLB-1 (without masking) achieved the top CS F1 score in its category, showing remarkable performance compared to off-the-shelf models and LLMs. NLLB-4 with Mask also obtained a competitive CS F1 score compared to those models equipped with masks, rivaling GPT-4 + Mask. In conclusion, these results underscore the importance of training models on source data rather than relying on off-the-shelf models. The NLLB results show that we achieve comparable outcomes to those of Google-NMT and Gemini-Pro on machine translation metrics, namely BLEU and chrF. On the other hand, the code-switch metric (CS F1) indicates that NLLB models retain medical keywords more effectively than off-the-shelf MT models.

\noindent
\textbf{Human preference}. As shown in Table~\ref{tab:human_eval_glicko}, human preference evaluation received responses from 23 medical doctors (MDs). The Glicko rating calculation results show that both NLLB models are preferred over Google NMT and LLMs like GPT-3.5. Both models are also almost equally preferred when compared to translations from Gemini-Pro models. Thus, we can summarize that machine translation metrics might not fully satisfy medical doctors' preferences. The results from the MT metrics contradict human preferences, which we will discuss further in Section~\ref{sec:fact_vs_mach}. Confirming our hypothesis,  we also found that MDs preferred CS translation over translating all words into the target language, as indicated by the CS F1 metric. MD preferences are discussed in more detail in the following section, Section~\ref{sec:discuss_md_eval}.

\begin{table}[!ht]
    \centering
    \caption{Models sorted by their Glicko ratings with 95\% confidence interval. Our fine-tuned NLLB models' scores are highlighted below} 
    \label{tab:human_eval_glicko} %
    \begin{footnotesize}
        \begin{tabular}{ll}
    \hline
       Model & Glicko MD  \\ \hline
        Human Annotated & 1638.57 $\pm$ 49.39 \\
        Gemini-CS & 1500.00 $\pm$ 50.31 \\ 
        Google NMT & 1398.61 $\pm$ 50.07 \\ 
        GPT-3.5-MN & 1316.40 $\pm$ 48.52 \\ 
        GPT-4-CS & 1578.93 $\pm$ 49.84 \\
        NLLB-1 & \textbf{1549.55} $\pm$ 52.05 \\ 
        \hline
        Gemini-MN + Mask & 1555.71 $\pm$ 55.18 \\ 
        Google NMT + Mask & 1480.98 $\pm$ 53.12 \\ 
        GPT-3.5-CS + Mask & 1394.69 $\pm$ 51.03 \\ 
        GPT-4-CS + Mask & 1564.60 $\pm$ 48.10 \\ 
        NLLB-3 + Mask & \textbf{1532.28} $\pm$ 50.79 \\
        \hline
    \end{tabular}
    \end{footnotesize}
\end{table}

\section{Discussion}


\subsection{Automated Metrics Versus Factual Accuracy}
\label{sec:fact_vs_mach}

The evaluation results reveal an unexpected outcome: Google NMT consistently achieves top scores across nearly all machine metrics among the 52 models, despite the lack of medical terminology preservation. Similarly, Google NMT with Mask dominates in almost every automated metric among the masked models (a better rank breakdown can be seen in Table~\ref{tab:rank_eval_result}). Nevertheless, a closer examination of individual samples still reveals that Google NMT frequently translates medical terminology imprecisely (as shown in Figure~\ref{img:example-cs-nllb}).
We hypothesize that Google NMT's superior performance in automated metrics is due to its fluency in translating non-essential parts of the medical text, which constitutes the bulk of our dataset. 
Conversely, the accuracy of medical-domain translations rather depends on the precise translation of critical medical terms, an area where Google NMT falls short. This is further supported by the minimal correlation between most automated metric scores and factual accuracy, especially among models that are rated higher than 3 in factual accuracy (see Figure ~\ref{fig:rate_vs_machine_zoom}).

In fact, the CS F1 metric addresses this issue by focusing on the preservation of key medical terms, demonstrating a stronger positive correlation with factual accuracy ratings. However, it is still not a comprehensive metric, as it only assesses the retention of English keywords without considering the quality of the Thai translation. A trade-off consideration between the retention of precise medical terms and the fluency of the overall translation may be necessary to develop a more suitable automated metric for medical translation tasks.

\subsection{MD Evaluation}
\label{sec:discuss_md_eval}

Our human evaluation within the MD population further supports our hypothesis that traditional automated metrics are not well-suited for medical-domain MT. This is shown by the significant correlation observed between Glicko ratings and both factual accuracy scores ($r = 0.698$) and CS F1 scores ($r = 0.516$), as opposed to the weak correlation (less than $0.3$) between traditional automated metrics and Glicko ratings (seen in Figure~\ref{fig:glicko_vs_metrics}).

\begin{figure}[!ht]
\vspace{-8mm}
    \centering
    \includegraphics[width=8cm]{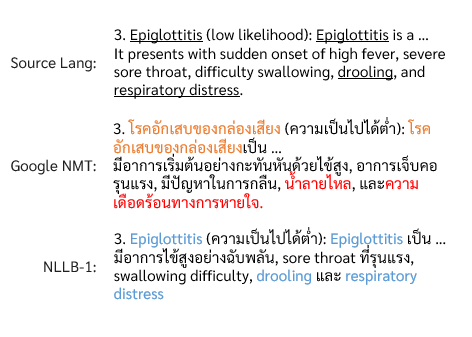}
    \vspace{-10mm}
    \caption{Real samples where our internal MDs and external MDs both report a preference for NLLB-1 CS translation over Google NMT. \textcolor{red}{Red sections} indicate medical keywords that Google NMT does not translate precisely. \textcolor{orange}{Orange sections} indicate medical keywords that Google NMT translates precisely, but retaining them in English is still preferred. \textcolor{blue}{Blue sections} indicate medical keywords that are retained in English and convey their meaning precisely.}
    \label{img:example-cs-nllb}
    \vspace{-5mm}
\end{figure}

Moreover, an in-depth analysis of questionnaire responses (shown in Figure~\ref{img:example-cs-nllb}) also presents a consistent picture. Google NMT provides a fluent Thai translation of the English text, but medical terminologies are still imprecisely translated. On the other hand, our NLLB model, despite exhibiting less fluency, successfully retains critical medical terminology in English. This also aligns with our hypothesis that traditional automated metrics tend to measure the fluency of the translation but not the precision of medical terminology translation. Therefore, in medical-domain translations, traditional automated metrics might not be adequate for measuring the quality of translations.

\section{Conclusion}
\label{sec:conclusion}

This paper presents an approach for performing MT in the medical domain using a CS translation to generate translations preferred by medical professionals. 
We developed a method for generating CS translation data, trained a CS translation model leveraging this data, and evaluated its performance against multiple strong baselines. 
The experimental results demonstrate that although most automated metrics might be suitable for measuring translation fluency, they are inadequate for assessing factual accuracy or medical doctors' preference in the translations. 
While current MT technologies may offer monolingual translations with high fluency, medical professionals exhibit a clear preference for CS medical translations that accurately preserve crucial terms in English, even at the expense of fluency. 
%

\section*{Limitations}
There are inherent risks associated with machine translation (MT), particularly the potential for misinterpreting medical terminology and technical terms. While our models have shown promising results, there is still a possibility of inaccuracies in translation that could affect daily practice.

Moreover, there is still some potential for further improvement. We have not conducted extensive human preference evaluations on all 52 models because doing so would require more than 390 MD respondents, whom we cannot find or hire. Also, we have not optimized prompts for LLMs to produce the best CS translations yet. Our MDs deemed the translations generated by these prompts acceptable internally, so we selected them. Lastly, we have not conducted an extensive hyperparameter search for NLLB training. To limit the cost of the fine-tuning process, we selected the standard learning rate and learning rate scheduler that is used throughout the field and fixed it for the entire fine-tuning process of NLLB.

\section*{Ethical Considerations}


Our human annotators were undergraduate students majoring in linguistics at a university in Thailand. We ensured that they received monthly monetary compensation at an industry-competitive salary. We compensated our annotators by first measuring their annotation speed in terms of the number of words processed per hour. After that, we established a monthly target for the annotators to achieve, and we paid our annotators a fixed salary.

Other human evaluators who respond to our questionnaire participate voluntarily. The participants were promised free usage of our upcoming product as compensation by randomly selecting five participants. In this regard, we have to collect their names and emails to prevent spamming attempts. For the remaining participants, we informed them that we would compensate for their work by releasing a questionnaire dataset without the respondents' information to the public domain, which we will release under a CC-BY-NC 4.0 license.

Regarding the licensing of models, we strictly adhere to the intended uses outlined by their respective licenses. The NLLB weight checkpoints we use as our pretrained weights are licensed under CC-BY-NC 4.0, which allows us to distribute our newly fine-tuned NLLB weights to the public for non-commercial use. We have also adhered to the Llama2 and SeaLLM Licenses by not using their outputs to enhance any language model and by restricting their use to research benchmarking purposes only. Additionally, we followed OpenAI and Google Gemini's Terms and Conditions strictly: we did not compete with OpenAI and Gemini’s models but rather used them fairly for research purposes.

All local LLM inferences, NLLB fine-tuning, and NLLB inferences for translation were performed on a single A100 GPU, also with the maximum amount of batching possible. We used a total of 60 GPU hours for fine-tuning NLLB, 24 GPU hours for performing local LLM inferences, and 3.5 GPU hours for performing 24 variants of NLLB inferences.

Regarding a potential leak of personal information, our source English texts inherently contain no personal information, as they are outputs from our own LLM product with no personal information in the prompt. We conduct an initial screening of the test benchmark dataset regardless, which confirms the absence of personal information in any of the English texts. We also instructed our internal annotator to remove any identifying information in case any is found within the annotation process. Another potential concern arises when we collect names and email addresses from MD evaluators. However, we use this information solely for spam tracking purposes and do not disclose or utilize this personal data for any other reason, except to contact individuals later regarding compensations for free usage.

\bibliography{custom}

\appendix

\label{sec:appendix}
\section{Fully listed Model Categories}
\label{sec:model_category}
\begin{compactitem}
    \item Google-translate-based model with keyword masking
    \item NLLB-based model with keyword masking 
    \item OpenThaiGPT-7B-based model with keyword masking
    \item OpenThaiGPT-13B-based model with keyword masking
    \item Typhoon-7B-based model with keyword masking
    \item SeaLLM-7B-based model with keyword masking
    \item LLama2-7B-based model with keyword masking
    \item LLama2-13B-based model with keyword masking
    \item Gemini-Pro-based model with keyword masking
    \item GPT-3.5-based model with keyword masking
    \item GPT-4-based model with keyword masking
    \item Google-translate-based model without keyword masking
    \item NLLB-based model without keyword masking
    \item OpenThaiGPT-7B-based model without keyword masking
    \item OpenThaiGPT-13B-based model without keyword masking
    \item Typhoon-7B-based model without keyword masking
    \item SeaLLM-7B-based model without keyword masking
    \item LLama2-7B-based model without keyword masking
    \item LLama2-13B-based model without keyword masking
    \item Gemini-Pro-based model without keyword masking
    \item GPT-3.5-based model without keyword masking
    \item GPT-4-based model without keyword masking
\end{compactitem}

\section{Prompts for Large Language  Model Translation}
\subsection*{CS Translation Prompt}
\begin{small}\begin{verbatim}
You are a linguist with expertise in medicine
and had your training in Thailand.
You are well acquainted to how's Thai MD
usually code switched between Thai Language and
English when they're communicating medical-related 
information among each other.
For instance, you never translate the following
English medical terms and jargons, symptoms,
technical terms, and pharmaceutical terms into Thai.

Hence, task is to examine the medical-related 
information text input and translated them
into Thai with the previously given constraint
and information.

\end{verbatim}\end{small}

\subsection*{Monolingual Translation Prompt}
\begin{small}\begin{verbatim}
Translate the following text input into Thai
in Medical Context
\end{verbatim}\end{small}

\subsection*{GPT4 Medical NER Prompt}
\begin{small}\begin{verbatim}
Annotate the medical report with HTML-like tags.
The output should start with <annotated> and end
with </annotated>.
Use the following tags to annotate the
respective terms:
- <patho> for pathological and medical symptoms
terms
- <pharm> for pharmaceutical terms and drugs' names
- <taxo> for scientific names and taxonomical-
like names
- <anato> for anatomical terms
- <chem> for chemical names
- <med> for medical practices and jargons
FYI:
- Drug names sometimes start with a single charactor
followed by full stop then full name.
For example: A. Parafivir, B. Paracetamol.
- Anatomical terms must include limbs, organs, cells,
and organelle.
\end{verbatim}\end{small}

\section{NLLB Training Configuration}
\label{sec:train_config}
\begin{small}\begin{verbatim}
LoraConfig:
    r = 16,
    lora_alpha = 16,
    target_modules = ["q_proj", "v_proj"],
    lora_dropout = 0.1,
    bias = "none",


TrainingArguments:
    num_train_epochs = 10,
    evaluation_strategy = "steps",
    logging_strategy ="steps",
    save_strategy ="steps",
    eval_steps=5000,
    logging_steps=500,
    save_steps=5000,
    
    bf16=True,

    seed=42,
    data_seed=42,

    warmup_ratio = 0.1,  
    learning_rate=10e-5,  

    per_device_train_batch_size= 3,
    per_device_eval_batch_size= 4,

    load_best_model_at_end=True,
    metric_for_best_model="loss",


\end{verbatim}\end{small}

\section{Evaluation Metric Implementation details}
\label{sec:eval_metric_impl}

The evaluation environment was established using Python 3.11, incorporating the following key libraries and their respective versions:

\begin{compactitem}
    \item \textbf{PyTorch 2.2.0:} Used for neural network-based computations and model loading, supporting the latest deep learning model features and optimizations.
    \item \textbf{NLTK 3.8.1:} Provided tools for text processing and evaluation metrics, including BLEU, METEOR, and CHRF scores.
    \item \textbf{PyThaiNLP 4.0.2:} Essential for processing the Thai language, used specifically for tokenizing Thai text and for the implementation of the NewMM tokenizer~\cite{phatthiyaphaibun2023pythainlp}.
    \item \textbf{JiWER 3.0.2:} Employed for calculating Word Error Rate (WER) and Character Error Rate (CER), key in assessing model performance in speech recognition tasks.
    \item \textbf{Unbabel Comet 2.2.1:} Employed for calculating the COMET score using the XCOMET-XL~\cite{guerreiro2023xcomet} model.
\end{compactitem}

We implemented a Python script on our own to calculate the Glicko rating based on \cite{glickman_parameter_1999}. The RD/Glicko evaluation was established using an initial rating of 1500 and an RD of 350. All the ratings are calculated at once, eliminating the need for nondeterminism. The 95\% confidence interval is reported using 2 times the RD.

\begin{table*} \begin{center}
    \caption{Internal Evaluation metric}\label{tab:band_score} %
    \begin{small}
    \begin{tabular}{ l|p{5.5cm}|p{5.5cm}}
     \hline
    \textbf{Band Score}	& \multicolumn{2}{c}{\textbf{Factual Correctness}} \\
    \hline
    7 &	\multicolumn{2}{c}{Fully contain all information, no addition or loses of information.}\\
    \hline
    6 & Fully contain all information but might add some information that does not improve (increased in clarity) of the source text. 
    	& Fully contain all information and also correctly adding information that enhanced the source text.\\
    \hline
    5	& Fully contain all information but might have some hallucination added in the translation but does not distort the information.& 
    	Losses of information that can safely disregarded.\\
    \hline
    4	& Fully contain all information but have some hallucination added and minorly distort information.
    	& Losses of information in such a way that might distort the information if the reader does not pay attention.\\
    \hline
    3	& Fully contain all information but have a noticeable amount of hallucination that majorly distort the information.
    	 & Losses of information that majorly distort the information in such a way that misled the reader.\\
    \hline
    2	& The reader can barely gain information from the translation.
    	& Hallucination heavily distorted the information that led to misunderstanding by the reader.\\
    \hline
    1	& The reader cannot gain the information from the translation.
    	& The translation not relevant to the source text.\\
    \hline
    \end{tabular}
    \end{small}
\end{center}\end{table*}

\section{Disclaimer for Participants}
\textbf{Notice to Participants}
\begin{compactitem}
\item This study focuses exclusively on medical questions.
\item Our system leverages LLM technology currently under development. Do not use the output as medical facts.
\item Participant's inputs, system outputs and feedback will be reviewed and used to improve the system capability.
\item To comply with Thai PDPA law, do not disclose real patient information or any patient identifiable information in general. Use hypothetical clinical cases only.
\end{compactitem}

\begin{figure*}
    \centering
\includegraphics[width=.55\textwidth]{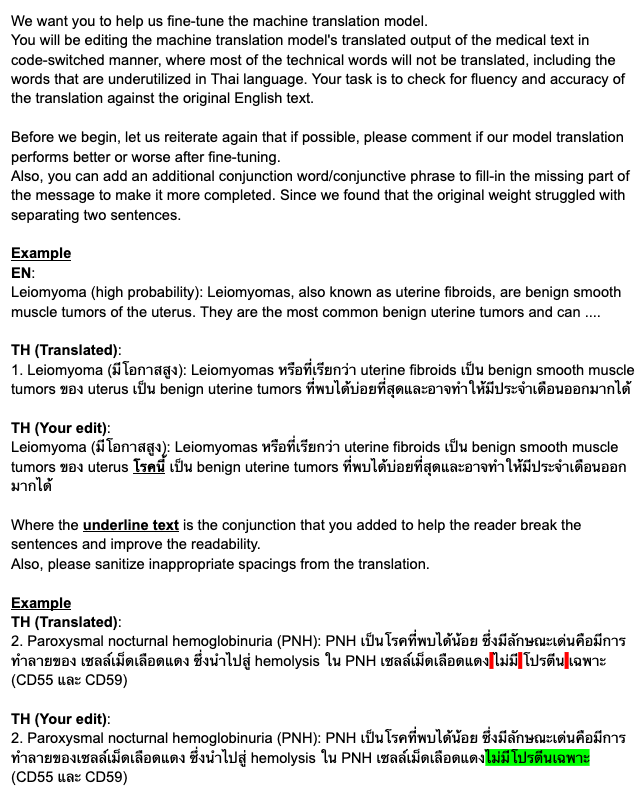}
    \caption{Instruction text for human annotators}
    \label{fig:annot-instruct}
\end{figure*}

\section{GPT-4's Medical NER Performance}
\label{sec:gpt4-ner-perf}
Although our confidence in GPT-4's capabilities in medical keyword extraction was already substantial, based on its performance in various analyses~\cite{nori2023generalist}, we have conducted an experiment to determine the medical NER performance of multiple systems. Results are shown in Table~\ref{tab:med-ner-results}

\begin{table}[H]
    \centering
    \caption{Performance of multiple LLMs and tools used to perform medical NER. CS F1 refers to the F1 score in identifying medical keywords denoted as "CS F1".}
    \label{tab:med-ner-results}
    \begin{tabular}{llll}
    \textbf{Model}      & \textbf{CS F1} & \textbf{Recall} & \textbf{Precision} \\
    \hline
    GPT-4      & 0.30 & 0.49 & 0.25         \\
    GPT-3.5    & 0.29 & 0.47 & 0.23         \\
    Gemini-Pro & 0.28 & 0.40 & 0.25         \\
    BioMedNER  & 0.03 & 0.05 & 0.03
    \end{tabular}
\end{table}

The complete evaluation of GPT-4 reveals scores of 0.488 and 0.253 for average CS recall and average CS precision, respectively, making it the top system for low-resource languages. Among the options considered, GPT-4 stands out as a competent system, particularly in its ability to detect keywords akin to how medical doctors code-switch in Thai. Although the initial mask detection is not perfect, we asked MDs to review our test dataset. If a mask is missing or incorrect, we request the MD to examine and correct it for us. Consequently, the test dataset is accurately masked.

\section{Medical Doctor Annotator Instruction}
\textbf{Instruction}
\begin{compactitem}
\item Please review the annotations from human annotators in the Google Spreadsheet provided.
\item Look for any serious errors in the labels.
\item Pay attention to whether any technical words have been lost during the cleaning process.
\item Leave a comment in the spreadsheet for any errors that you may find.
\end{compactitem}

\begin{table*}[!ht]
    \centering
    \caption{Factual Evaluation Result per MD} 
    \label{tab:internal_eval_result_full} %
    \begin{small}
     \begin{tabular}{llllll}
    \hline
       Model Variant & A* & B* & C* & D* & Fact. Score \\ \hline
        Gemini-Pro-CS & 4.5 & 6.5 & 5 & 7 & 5.75 \\ 
        Gemini-Pro-MN & 6.5 & 6.5 & 3.5 & 5 & 5.375 \\ 
        Google-NMT & 3 & 4 & 3.5 & 2 & 3.125 \\ 
        GPT-3.5-CS & 3 & 5 & 3 & 3.5 & 3.625 \\ 
        GPT-3.5-MN & 3 & 5 & 3.5 & 4 & 3.875 \\ 
        GPT-4-CS & 5 & 7 & 6 & 7 & 6.25 \\ 
        GPT-4-MN & 4.5 & 5.5 & 5 & 5.5 & 5.125 \\ 
        Llama2-13B-CS & 1 & 1.5 & 0 & 2 & 1.125 \\ 
        Llama2-13B-MN & 1 & 1 & 0 & 2 & 1 \\ 
        Llama2-7B-CS & 1 & 0 & 0 & 1 & 0.5 \\ 
        Llama2-7B-MN & 1 & 0 & 0 & 1 & 0.5 \\ 
        OpenThaiGPT-13B-CS & 2 & 1 & 3 & 2.5 & 2.125 \\ 
        OpenThaiGPT-13B-MN & 2.5 & 3 & 1.5 & 2.5 & 2.375 \\ 
        OpenThaiGPT-7B-CS & 2 & 4 & 2 & 3.5 & 2.875 \\ 
        OpenThaiGPT-7B-MN & 2.5 & 4.5 & 1 & 3 & 2.75 \\ 
        SeaLLM-7B-CS & 2 & 3 & 1 & 2 & 2 \\ 
        SeaLLM-7B-MN & 2 & 2.5 & 0 & 2.5 & 1.75 \\ 
        Typhoon-7B-CS & 1.5 & 2 & 2 & 2 & 1.875 \\ 
        Typhoon-7B-MN & 1.5 & 1 & 2 & 3 & 1.875 \\ 
        NLLB & 1.5 & 3.5 & 1.5 & 3.5 & 2.5 \\ 
        NLLB-1 & 3.5 & 5 & 4.5 & 4.5 & 4.375 \\ 
        NLLB-2 & 3 & 4 & 2.5 & 4 & 3.375 \\ 
        NLLB-3 & 3.5 & 5.5 & 3.5 & 3.5 & 4 \\ 
        NLLB-4 & 2.5 & 5 & 3.5 & 4.5 & 3.875 \\ 
        NLLB-5 & 3 & 4.5 & 2.5 & 3 & 3.25 \\ 
        NLLB-6 & 3.5 & 5.5 & 3.5 & 3.5 & 4 \\ \hline
        Gemini-Pro-CS + Mask & 5 & 6.5 & 5 & 5.5 & 5.5 \\ 
        Gemini-Pro-MN + Mask & 5 & 7 & 4.5 & 6.5 & 5.75 \\ 
        Google-NMT + Mask & 4 & 6 & 4.5 & 5.5 & 5 \\ 
        GPT-3.5-CS + Mask & 5 & 7 & 4 & 5 & 5.25 \\ 
        GPT-3.5-MN + Mask & 5.5 & 5.5 & 4.5 & 4.5 & 5 \\ 
        GPT-4-CS + Mask & 5 & 6.5 & 6 & 6.5 & 6 \\ 
        GPT-4-MN + Mask & 3.5 & 6.5 & 3.5 & 5.5 & 4.75 \\ 
        Llama2-13B-CS + Mask & 3 & 3 & 1 & 2 & 1 \\ 
        Llama2-13B-MN + Mask & 1 & 0 & 0 & 2 & 0.75 \\ 
        Llama2-7B-CS + Mask & 1 & 0 & 0 & 1 & 0.5 \\ 
        Llama2-7B-MN + Mask & 1 & 1 & 0 & 1 & 0.75 \\ 
        OpenThaiGPT-13B-CS + Mask & 1 & 1 & 0 & 2 & 2.25 \\ 
        OpenThaiGPT-13B-MN + Mask & 2 & 1.5 & 0 & 2 & 1.375 \\ 
        OpenThaiGPT-7B-CS + Mask & 1.5 & 1 & 1 & 1.5 & 1.25 \\ 
        OpenThaiGPT-7B-MN + Mask & 1 & 4 & 0.5 & 3 & 2.125 \\ 
        SeaLLM-7B-CS + Mask & 1.5 & 2 & 0 & 2 & 1.375 \\ 
        SeaLLM-7B-MN + Mask & 1.5 & 2.5 & 0 & 2.5 & 1.625 \\ 
        Typhoon-7B-CS + Mask & 1.5 & 1 & 2 & 3 & 1.875 \\ 
        Typhoon-7B-MN + Mask & 1 & 1 & 2 & 3 & 1.75 \\ 
        NLLB + Mask & 3.5 & 6 & 2 & 5 & 4.125 \\ 
        NLLB-1 + Mask & 5 & 6 & 4.5 & 3 & 4.625 \\ 
        NLLB-2 + Mask & 3.5 & 6 & 4 & 5.5 & 4.75 \\ 
        NLLB-3 + Mask & 4 & 6.5 & 4 & 6 & 5.125 \\ 
        NLLB-4 + Mask & 3.5 & 6.5 & 4 & 5.5 & 4.875 \\ 
        NLLB-5 + Mask & 2 & 5 & 3 & 4.5 & 3.625 \\ 
        NLLB-6 + Mask & 4 & 6 & 3.5 & 5.5 & 4.75 \\ \hline
    \end{tabular}
    \end{small}
\end{table*}

\begin{figure*}
\centering
    \includegraphics[width=.23\textwidth]{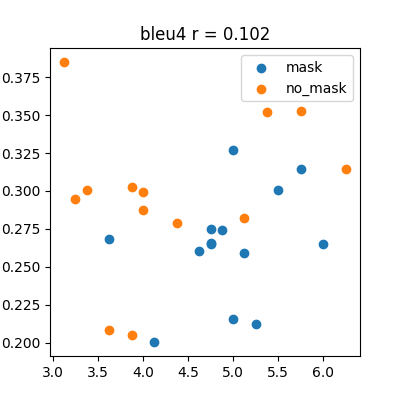}
     \includegraphics[width=.23\textwidth]{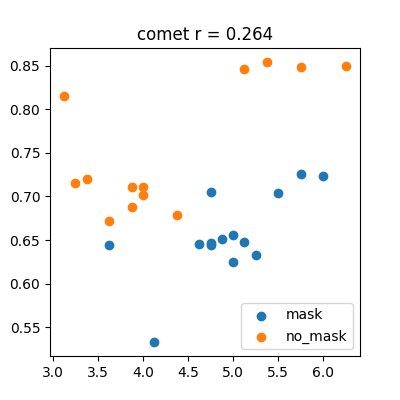}
    \includegraphics[width=.23\textwidth]{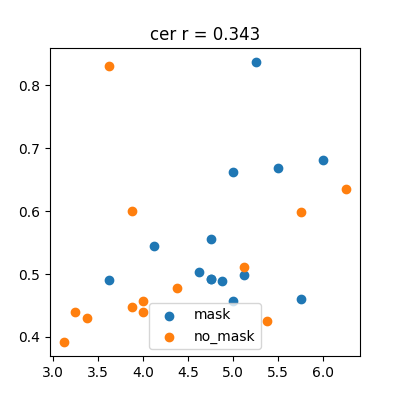}
    \includegraphics[width=.23\textwidth]{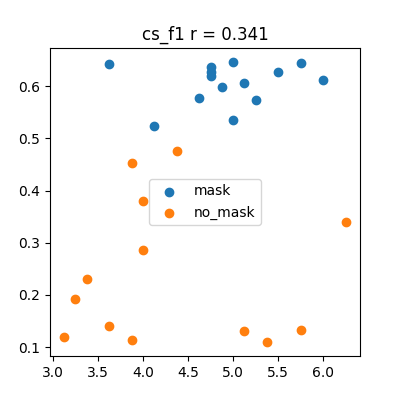}
    \\[\smallskipamount]
    \includegraphics[width=.23\textwidth]{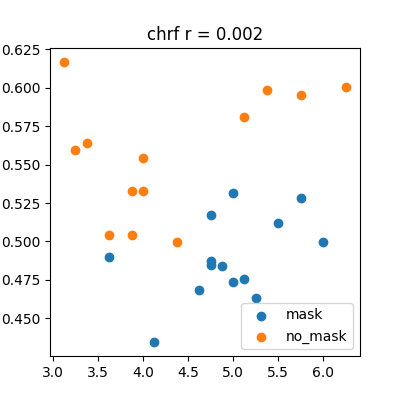}
    \includegraphics[width=.23\textwidth]{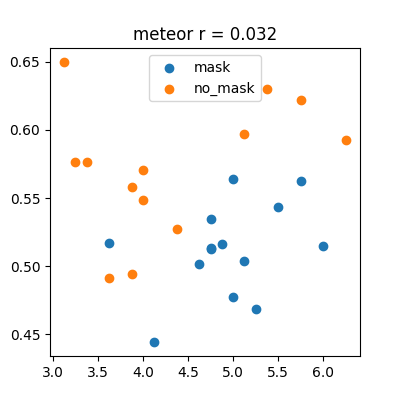}
    \includegraphics[width=.23\textwidth]{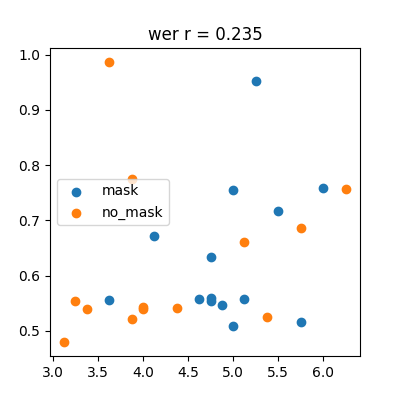}
    \\[\smallskipamount]

\caption{Plots of factual score of each model that pass 3 factual accuracy score against machine evaluation metric.Masked model are labeled in blue and models without masked are labeled in orange.}
\label{fig:rate_vs_machine_zoom}
\end{figure*}

\begin{figure*}
\centering

    \includegraphics[width=.23\textwidth]{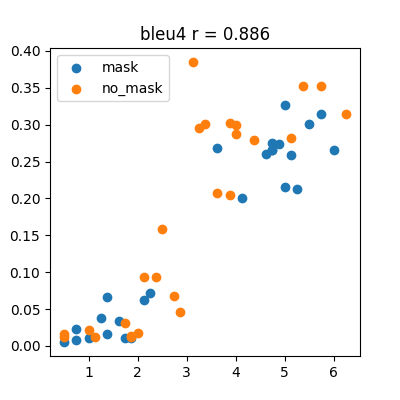}
     \includegraphics[width=.23\textwidth]{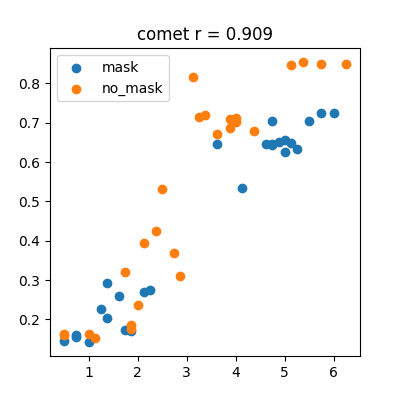}
    \includegraphics[width=.23\textwidth]{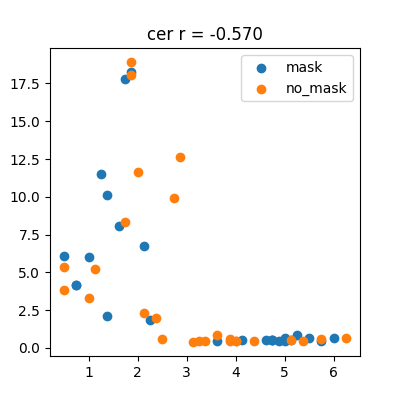}
    \includegraphics[width=.23\textwidth]{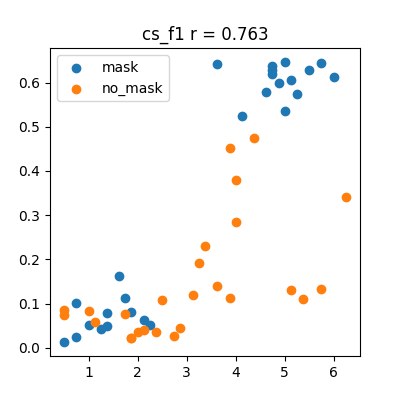}
    \\[\smallskipamount]
    \includegraphics[width=.23\textwidth]{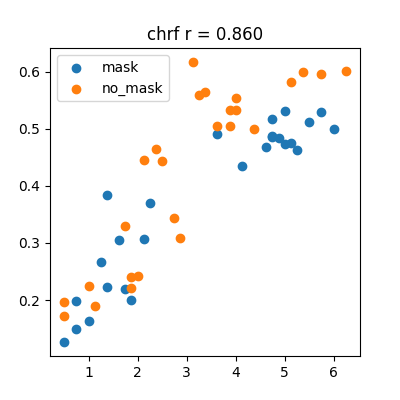}
    \includegraphics[width=.23\textwidth]{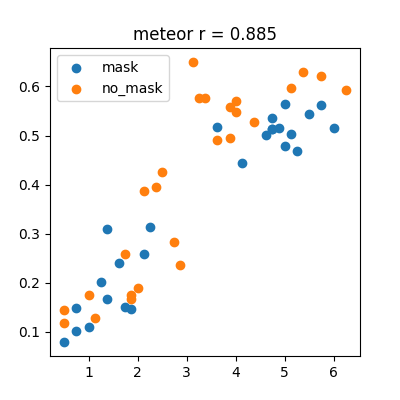}
    \includegraphics[width=.23\textwidth]{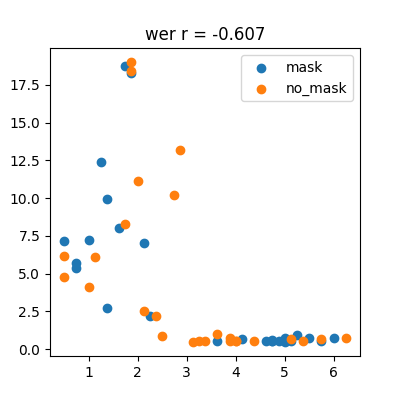}
    \\[\smallskipamount]
\caption{Plots of a factual score of all models against machine evaluation metric. The model below 3 factual accuracy score gives an extremely high value of CER and WER, so we decided to exclude them to not skew the Pearson r coefficient in the \ref{fig:rate_vs_machine_zoom}. }
\label{fig:rate_vs_machine}
\end{figure*}

\begin{figure*}
\centering

    \includegraphics[width=.23\textwidth]{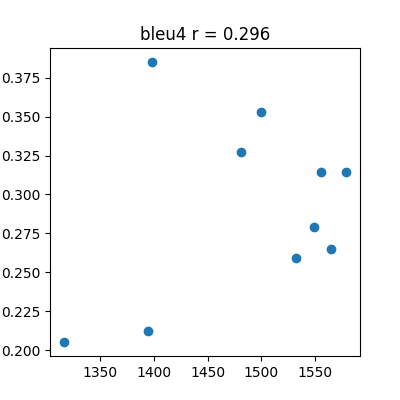}
     \includegraphics[width=.23\textwidth]{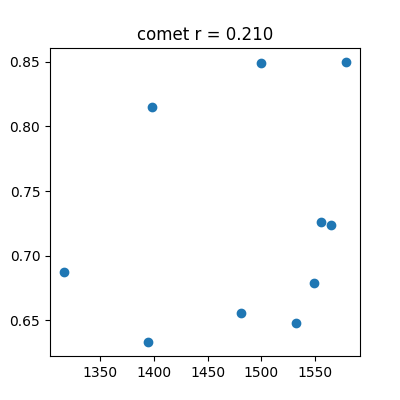}
    \includegraphics[width=.23\textwidth]{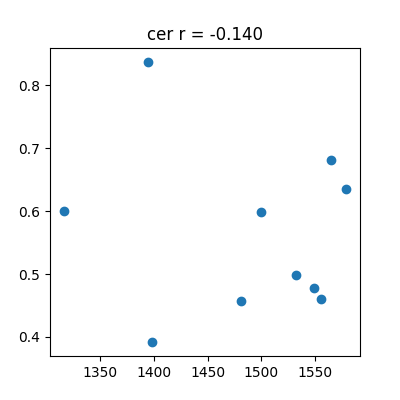}
    \includegraphics[width=.23\textwidth]{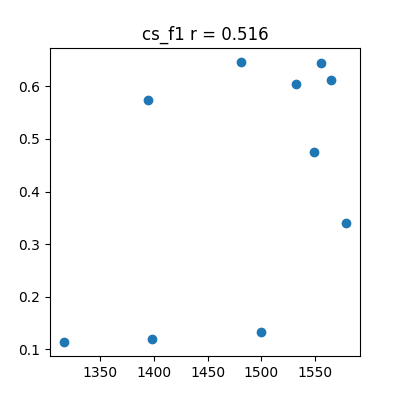}
    \\[\smallskipamount]
    \includegraphics[width=.23\textwidth]{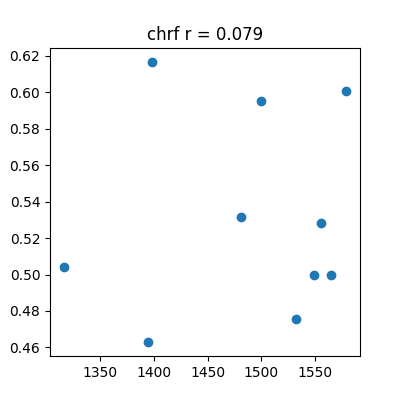}
    \includegraphics[width=.23\textwidth]{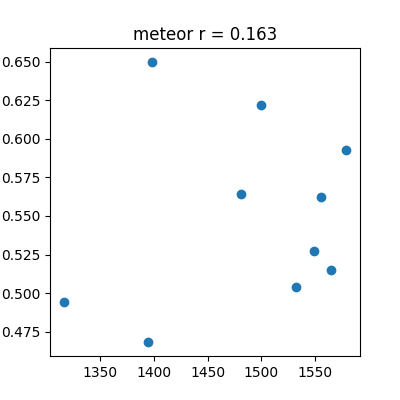}
    \includegraphics[width=.23\textwidth]{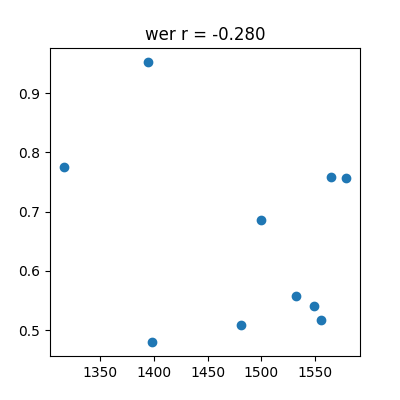}
    \includegraphics[width=.23\textwidth]{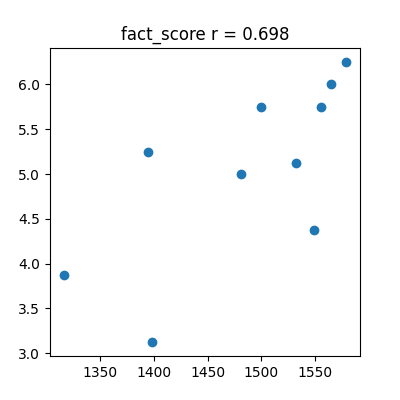}
    \\[\smallskipamount]
\caption{Plots of Glicko score of all human evaluated models against machine evaluation metric and Factual accuracy score.}
\label{fig:glicko_vs_metrics}
\end{figure*}

\begin{table*}[!ht]
    \centering
    \caption{Model's rank on each score type}
    \label{tab:rank_eval_result} %
    \begin{footnotesize}
   \begin{tabular}{l|lllllll|ll}
    \hline
       Rank of each score & CSF1 & BLEU & chrF & CER & WER & COMET & METEOR & Fact. \\ \hline
        Gemini-Pro-CS & 24.5 & 2 & 4 & 20.5 & 19 & 3 & 3 & 3.5 \\ 
        Gemini-Pro-MN & 29 & 3 & 3 & 2 & 4 & 1 & 2 & 6 \\ 
        Google-NMT & 26 & 1 & 1 & 1 & 1 & 5 & 1 & 27 \\ 
        GPT-3.5-CS & 23 & 25 & 15.5 & 27 & 28 & 17 & 20 & 23.5 \\ 
        GPT-3.5-MN & 27 & 26 & 15.5 & 20.5 & 25 & 15 & 17 & 21.5 \\ 
        GPT-4-CS & 18 & 5.5 & 2 & 23 & 23 & 2 & 5 & 1 \\ 
        GPT-4-MN & 24.5 & 8 & 5 & 15.5 & 18 & 4 & 4 & 8.5 \\ 
        Llama2-13B-CS & 39 & 46 & 48 & 37 & 37 & 50 & 48 & 45 \\ 
        Llama2-13B-MN & 33 & 40 & 41 & 33 & 33 & 45 & 41 & 46.5 \\ 
        Llama2-7B-CS & 37 & 46 & 49 & 38 & 38 & 48 & 49 & 51 \\ 
        Llama2-7B-MN & 32 & 43 & 47 & 34 & 34 & 46 & 47 & 51 \\ 
        OpenThaiGPT-13B-CS & 45 & 29.5 & 28 & 32 & 31 & 30 & 30 & 33.5 \\ 
        OpenThaiGPT-13B-MN & 46 & 29.5 & 24 & 30 & 29 & 29 & 29 & 31 \\ 
        OpenThaiGPT-7B-CS & 43 & 35 & 35 & 48 & 48 & 33 & 37 & 28 \\ 
        OpenThaiGPT-7B-MN & 48 & 32 & 33 & 44 & 45 & 31 & 33 & 29 \\ 
        SeaLLM-7B-CS & 47 & 41 & 39 & 47 & 46 & 38 & 39 & 35 \\ 
        SeaLLM-7B-MN & 36 & 38 & 34 & 43 & 43 & 32 & 34 & 39.5 \\ 
        Typhoon-7B-CS & 51 & 46 & 43 & 52 & 50 & 41 & 42 & 37 \\ 
        Typhoon-7B-MN & 50 & 44 & 40 & 50 & 52 & 42 & 40 & 37 \\ 
        NLLB & 30 & 28 & 29 & 22 & 26 & 28 & 28 & 30 \\ 
        NLLB-1 & 15 & 16 & 18 & 10 & 6 & 16 & 16 & 17 \\ 
        NLLB-2 & 20 & 13 & 6 & 3 & 14.5 & 8 & 8.5 & 25 \\ 
        NLLB-3 & 17 & 14 & 11.5 & 9 & 8 & 14 & 14 & 19.5 \\ 
        NLLB-4 & 16 & 10 & 11.5 & 7.5 & 5 & 11 & 13 & 21.5 \\ 
        NLLB-5 & 21 & 15 & 7 & 5.5 & 16 & 9 & 8.5 & 26 \\ 
        NLLB-6 & 19 & 12 & 8 & 4 & 11 & 10 & 11 & 19.5 \\ \hline
        Gemini-Pro-CS + Mask & 5.5 & 7 & 14 & 25 & 20 & 13 & 10 & 5 \\ 
        Gemini-Pro-MN + Mask & 2 & 5.5 & 10 & 7.5 & 3 & 6 & 7 & 3.5 \\ 
        Google-NMT + Mask & 1 & 4 & 9 & 5.5 & 2 & 18 & 6 & 10.5 \\ 
        GPT-3.5-CS + Mask & 12 & 24 & 26 & 28 & 27 & 25 & 26 & 7 \\ 
        GPT-3.5-MN + Mask & 13 & 23 & 21 & 24 & 22 & 26 & 25 & 10.5 \\ 
        GPT-4-CS + Mask & 8 & 11 & 17 & 26 & 24 & 7 & 15 & 2 \\ 
        GPT-4-MN + Mask & 7 & 9 & 13 & 18.5 & 17 & 12 & 12 & 14 \\ 
        Llama2-13B-CS + Mask & 40.5 & 48.5 & 50 & 39 & 41 & 52 & 50 & 46.5 \\ 
        Llama2-13B-MN + Mask & 31 & 39 & 45.5 & 36 & 35 & 49 & 45 & 48.5 \\ 
        Llama2-7B-CS + Mask & 52 & 52 & 52 & 40 & 40 & 51 & 52 & 51 \\ 
        Llama2-7B-MN + Mask & 49 & 51 & 51 & 35 & 36 & 47 & 51 & 48.5 \\ 
        OpenThaiGPT-13B-CS + Mask & 40.5 & 31 & 32 & 29 & 30 & 35 & 31 & 32 \\ 
        OpenThaiGPT-13B-MN + Mask & 35 & 33 & 31 & 31 & 32 & 34 & 32 & 42.5 \\ 
        OpenThaiGPT-7B-CS + Mask & 44 & 36 & 38 & 46 & 47 & 39 & 38 & 44 \\ 
        OpenThaiGPT-7B-MN + Mask & 38 & 34 & 36 & 41 & 39 & 36 & 35 & 33.5 \\ 
        SeaLLM-7B-CS + Mask & 42 & 42 & 42 & 45 & 44 & 40 & 43 & 42.5 \\ 
        SeaLLM-7B-MN + Mask & 22 & 37 & 37 & 42 & 42 & 37 & 36 & 41 \\ 
        Typhoon-7B-CS + Mask & 34 & 48.5 & 45.5 & 51 & 49 & 44 & 46 & 37 \\ 
        Typhoon-7B-MN + Mask & 28 & 50 & 44 & 49 & 51 & 43 & 44 & 39.5 \\
        NLLB + Mask & 14 & 27 & 30 & 18.5 & 21 & 27 & 27 & 18 \\ 
        NLLB-1 + Mask & 11 & 21.5 & 27 & 17 & 9.5 & 22.5 & 24 & 16 \\ 
        NLLB-2 + Mask & 4 & 20 & 20 & 14 & 14.5 & 24 & 21.5 & 14 \\ 
        NLLB-3 + Mask & 9 & 21.5 & 25 & 15.5 & 12 & 20 & 23 & 8.5 \\ 
        NLLB-4 + Mask & 10 & 17 & 23 & 11 & 7 & 19 & 18.5 & 12 \\ 
        NLLB-5 + Mask & 3 & 18 & 19 & 12 & 13 & 22.5 & 18.5 & 23.5 \\ 
        NLLB-6 + Mask & 5.5 & 19 & 22 & 13 & 9.5 & 21 & 21.5 & 14 \\ \hline
    \end{tabular}
    \end{footnotesize}
\end{table*}

\begin{figure*}
\centering
    \includegraphics[width=\textwidth]{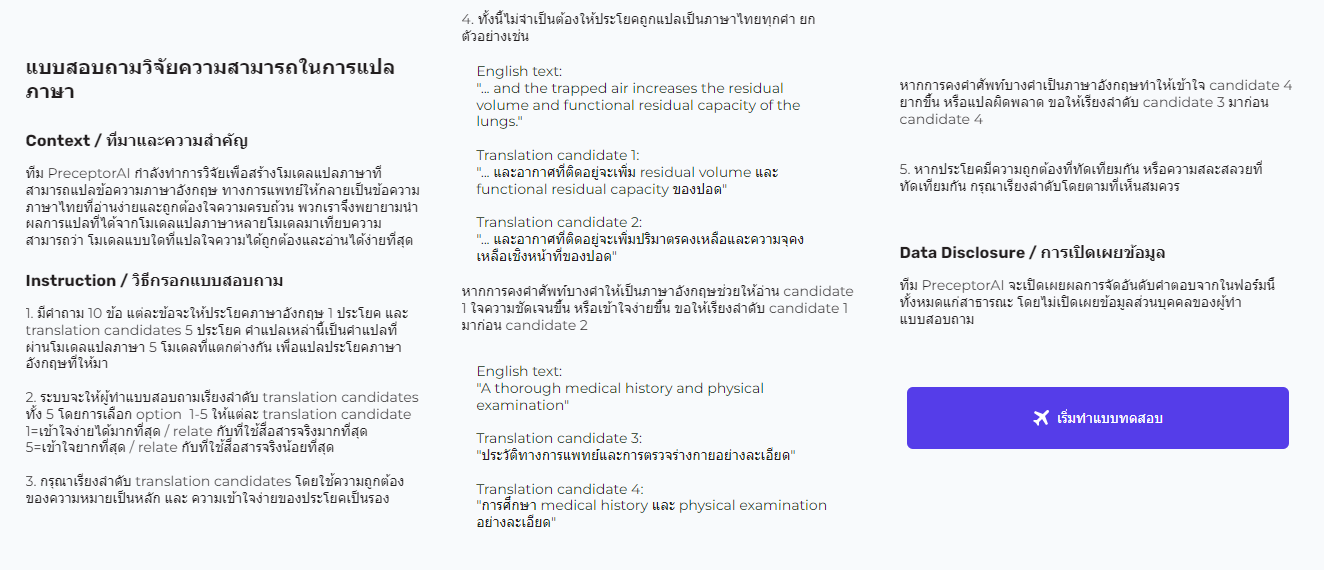}
\caption{Instruction page for respondents to respond to our questionnaire. }
\label{fig:questionnaire_sample0}
\end{figure*}

\end{document}